\pdfoutput=1

\documentclass[11pt]{article}

\usepackage[preprint]{acl}

\usepackage{times}
\usepackage{latexsym}

\usepackage[T1]{fontenc}

\usepackage[utf8]{inputenc}

\usepackage{microtype}

\usepackage{inconsolata}

\usepackage{graphicx}

\usepackage{tabularx}
\usepackage{algorithm}
\usepackage{algorithmic}
\usepackage{multirow}
\usepackage{colortbl}
\usepackage{booktabs}
\usepackage{makecell}
\usepackage{xcolor}
\usepackage{bbding}
\usepackage{pifont}
\usepackage{amsmath}

\title{Training LLM-Based Agents with Synthetic Self-Reflected Trajectories \\ and Partial Masking}

\author{
 \textbf{Yihan Chen\textsuperscript{1}},
 \textbf{Benfeng Xu\textsuperscript{1}},
 \textbf{Xiaorui Wang\textsuperscript{2}},
 \textbf{Yongdong Zhang}\textsuperscript{1}, 
 \textbf{Zhendong Mao\textsuperscript{1}},
\\
 \textsuperscript{1}University of Science and Technology of China,
 \textsuperscript{2}Metastone Technology
\\
\texttt{\{chenyihan, benfeng\}@mail.ustc.edu.cn}}

\begin{document}
\maketitle
\begin{abstract}
Autonomous agents, which perceive environments and take actions to achieve goals, have become increasingly feasible with the advancements in large language models (LLMs). However, current powerful agents often depend on sophisticated prompt engineering combined with closed-source LLMs like GPT-4. Although training open-source LLMs using expert trajectories from teacher models has yielded some improvements in agent capabilities, this approach still faces limitations such as performance plateauing and error propagation. 
To mitigate these challenges, we propose STeP, a novel method for improving LLM-based agent training. We synthesize \textit{self-reflected trajectories} that include reflections and corrections of error steps, which enhance the effectiveness of LLM agents in learning from teacher models, enabling them to become agents capable of self-reflecting and correcting. We also introduce \textit{partial masking} strategy that prevents the LLM from internalizing incorrect or suboptimal steps. 
Experiments demonstrate that our method improves agent performance across three representative tasks: ALFWorld, WebShop, and SciWorld. For the open-source model LLaMA2-7B-Chat, when trained using self-reflected trajectories constructed with Qwen1.5-110B-Chat as the teacher model, it achieves comprehensive improvements with less training data compared to agents trained exclusively on expert trajectories.
\end{abstract}

\section{Introduction}
The comprehensive capabilities of large language models (LLMs) have advanced significantly, enabling them to perform agent tasks that require interaction with various environments to reason, plan, and execute actions \cite{agentsurvey1, agentsurvey2}. Recently, many studies utilize GPT-4 \cite{openai2024gpt4technicalreport} or other powerful closed-source LLMs to build agents capable of executing complex tasks, such as manipulating computer systems \cite{xie2024osworld}, performing embodied intelligence tasks \cite{swiftsage}, and even resolving issues in real GitHub repositories \cite{yang2024sweagent}. These tasks are often accomplished through sophisticated prompting engineering. However, open-source models and smaller LLMs generally underperform in agent tasks compared to their closed-source counterparts \cite{liu2024agentbench, mialon2024gaia, xie2024travelplanner}.

\begin{figure}[t]
\centering
\includegraphics[width=1\columnwidth]{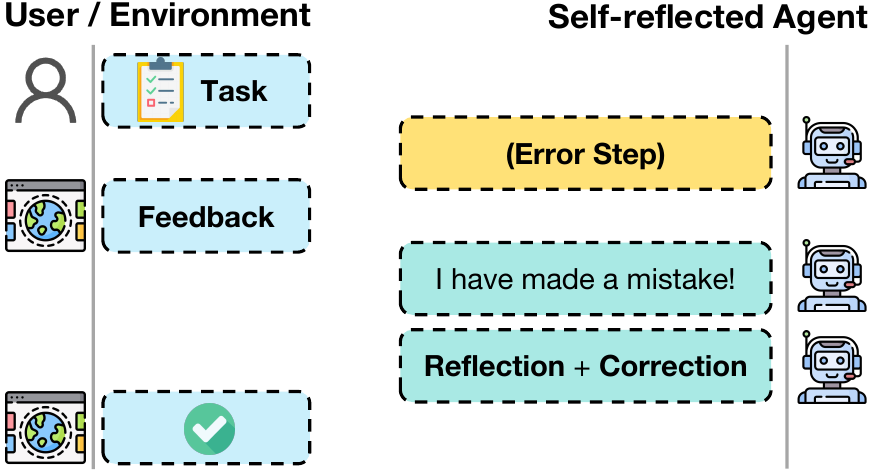} 
\caption{A self-reflected agent could autonomously identify, reflect on and correct errors based on interaction history.}
\label{intro}
\end{figure}

Previous works on building agents with small open-source LLMs often rely on supervised fine-tuning (SFT) \cite{agenttuning, chen2023fireact}. Due to the limited instruction-following and reasoning abilities of small LLMs \cite{ifeval}, these studies typically resort to naive distillation, employing prompting techniques like ReAct \cite{yao2023react} to guide the teacher model (GPT-4) in solving problems step by step while collecting their interaction trajectories with the environment. These trajectories are filtered for correct problem-solving instances and used as multi-turn dialogue data to SFT small LLMs.
However, training via naive distillation, where the small model mimics expert trajectories from the teacher, faces challenges. 
First, as the training data grows, performance improvements slow down and may eventually plateau. This indicates such methods have limitations in enhancing generalization. 
Second, LLM agents are prone to cascading errors: a single mistake can propagate further failures, trapping the agent in error loops \cite{qin2024toolllm}. 
Moreover, it remains unclear how to design an agent with self-reflective capabilities akin to proprietary models like GPT-4.

\begin{figure}[t]
\centering
\includegraphics[width=1\columnwidth]{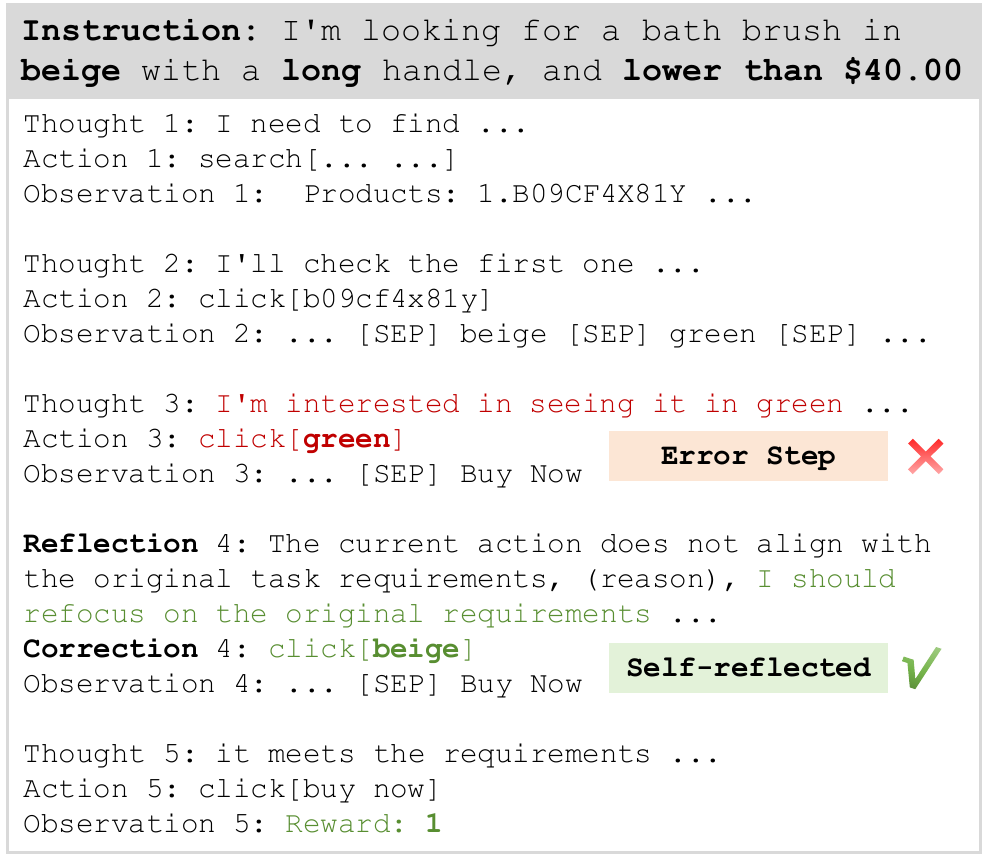}
\caption{\textit{Self-Reflected Trajectories} on WebShop.}
\label{example}
\end{figure}

Inspired by this, we propose \textbf{STeP} (\textbf{S}elf-Reflected \textbf{T}raj\textbf{e}ctories and \textbf{P}arital Masking), a novel method that enables small LLMs to learn to self-reflect and self-correct while interacting with the environment (Figure \ref{intro}). This approach could improve the effectiveness and efficiency of learning agent capabilities from teacher LLMs. STeP augments the training dataset by incorporating \textbf{\textit{Self-Reflected Trajectories}} that involve reflection on and correction of error steps. An example of them is depicted in Figure \ref{example}. During training, we introduce \textbf{\textit{Partial Masking}}, a novel masking mechanism designed to prevent the LLM from internalizing incorrect thoughts and actions. 

Specifically, our approach begins by fine-tuning an LLM on a subset of successful expert trajectories, creating a base LLM agent. For the remaining data, we introduce a stronger LLM as a teacher model to evaluate in real-time whether the actions taken by the base LLM agent are correct. If an action is deemed incorrect, the teacher model provides reflection and correction to guide the base agent back onto the correct trajectory. To ensure data consistency, all trajectories are converted into the ReAct format \cite{yao2023react}. Finally, we collect trajectories that include reflection and correction steps while completing tasks, then incorporate them into the training set to retrain the open-sourced LLM with \textit{Partial Masking} to further enhance its agent ability.

We perform our method on three representative tasks: ALFWorld \cite{alfworld} (simulated daily house-holding), WebShop \cite{webshop} (simulated online shopping), and SciWorld \cite{sciworld} (simulated science experiments). Initially, we train LLaMA2-7B-chat \cite{touvron2023llama2} to obtain a base LLM agent.
Then, we employ the LLM teacher to acquire self-reflected trajectories. We found that without utilizing a more costly closed-source LLM, we achieved improvements under the guidance of a larger-scale open-sourced LLM teacher.
Our results illustrate when utilizing Qwen1.5-110B-Chat \cite{bai2023qwentechnicalreport} as the teacher model, LLaMA2-7B-chat trained with self-reflected trajectories achieves comprehensive improvements compared to the model trained with the full dataset of expert trajectories. Additionally, we conduct ablation experiments to verify the necessity of \textit{Self-Reflected Trajectories} and \textit{Partial Masking}, as well as test the performance of existing models on these agent tasks.
The main contributions of this paper can be summarized as:
\begin{itemize}
\item We propose STeP, a method aiming to improve agent training that utilizes \textit{Self-Reflected Trajectories}, enabling the LLM to learn more effectively from the LLM teacher.
\item We introduce the \textit{Partial Masking} strategy, which helps prevent the LLM from learning erroneous or suboptimal steps in multi-turn trajectories.
\item Experiments on representative agent tasks confirm the efficacy of our approach. Our work emphasizes the importance of reflecting on and correcting errors in agent tasks.
\end{itemize}

\section{Related Works}
\paragraph{LLM-based Agent}
An autonomous agent is a system embedded in and interacting with an environment, possessing the ability to sense and act upon it \cite{AUTONOMOUSAGENTS}. The development of autonomous agents is widely viewed as a promising avenue toward achieving artificial general intelligence (AGI) \cite{agentsurvey1}. Currently, large language models (LLMs) are particularly well-suited for constructing these agents. By using prompting methods \cite{wei2023chainofthought, yao2023react} or more complex prompting strategies \cite{qian2024ice}, closed-source LLMs can effectively function as controllers to leverage tools, solve problems, write code, and complete tasks on real-world websites, etc. \cite{patil2023gorilla, opendevin2024, yang2024sweagent, gur2024webagent}.

There are also several works focusing on training open-source LLMs to enhance their capabilities as agents. FireAct \cite{chen2023fireact} and AgentTuning \cite{agenttuning} leverage pre-constructed successful expert trajectories from teacher LLM as multi-turn conversation data to fine-tune LLM, which is also referred to as behavioral cloning \cite{bc}. Additionally, some research employs multi-agent systems \cite{qiao2024autoact} or policy gradient optimization \cite{yao2024retroformer, xi2024agentgym} for training LLM-based agents. Furthermore, certain research tries to steer the model away from negative samples \cite{wang2024nat, song2024eto}. However, negative trajectories may still contain correct steps. Avoiding the entire trajectory consequently means avoiding the correct actions in the earlier parts. Unlike the approaches that simply use negative samples, our work offers a more granular approach, allowing for better distinguishing between correct and incorrect steps. 

\paragraph{Reflection in LLMs}
Reflection mechanisms utilize feedback from the environment or models to reflect on previous actions taken by LLMs, thereby enhancing their existing reasoning trace and task-specific action choice abilities. Currently, this approach has widespread applications in the field of developing more capable LLMs \cite{peng2023checkfactstryagain, shinn2023reflexion, madaan2023selfrefine, gupta2024metareflection}.
Previous approaches usually employ heuristic triggering conditions that activate reflection only when the action repetition exceeds a limit or when the trajectory reaches a maximum step count without completing the task. However, if LLM-based agents have limited context length, reaching the maximum step count may leave insufficient context to complete the task. Moreover, heuristic detection methods can only address a limited range of error types and are unable to detect and correct arbitrary errors. 
Compared to directly utilizing prior Reflection-related works such as Reflexion \cite{shinn2023reflexion} for trajectory construction, we employ real-time evaluation, making it more flexible and less demanding on contextual space.

\section{Method}
In this section, we first introduce the background knowledge, followed by a detailed description of the three stages of STeP: Agent Initialization, \textit{Self-Reflected Trajectories} Synthesizing, and SFT with \textit{Partial Masking}.

\subsection{Preliminaries}
Given the agent task that contains the interaction with an environment, an LLM-based agent is required to generate actions based on the task instruction and continuously advance or adjust those actions according to feedback from the environment until the task is completed. The interaction trajectory between the LLM-based agent and the user environment $e$ can be described as a multi-turn conversation history \cite{agenttuning}. The agent task can also be formalized as a partially observable Markov decision process: $(\mathcal{U}, \mathcal{S}, \mathcal{A}, \mathcal{O}, \mathcal{T}, \mathcal{R})$ \cite{song2024eto, xi2024agentgym}. The elements within it represent the task instruction space, state space, action space, observation space, state transition function, and reward function, respectively.

For each instruction $u \in \mathcal{U}$, the LLM-based agent $\pi_{\theta}$ will generate the action $a_1 \in \mathcal{A}$, Under the influence of $a_1$, the initial task latent state $s_0 \in \mathcal{S}$ transitions to $s_1 \in \mathcal{S}$ according to $\mathcal{T}$. Meanwhile, the environment $e$ will provide an observation in response to $a_1$, yielding feedback $o_1 \in \mathcal{O}$.
The subsequent action $a_i (i >= 2)$ will be generated based on the instruction $u$, previous actions and environment feedback, i.e. $a_i = \pi_{\theta} (u, a_1, o_1, ..., a_{i-1}, o_{i-1})$.
The environment $e$ will provide the corresponding observation $o_i$ based on the transition in the latent state space due to $a_i$. This process will iterate until the task is completed or the maximum step limit is reached. We denote the number of actions at this point as the trajectory length $n$. The trajectory $\tau$ is represented as:
\begin{equation*}
\tau = (u, a_1, o_1, ..., a_{i}, o_{i}, ..., a_{n}, o_{n})
\end{equation*}
\begin{equation*}
\pi_{\theta} (\tau | e, u) = \prod_{i=1}^{n} \pi_{\theta} (a_i|e, u, a_1, o_1, ..., a_{i-1}, o_{i-1})
\end{equation*}

Once the task is ended, the reward $r \in [0, 1]$ will be calculated based on reward function $\mathcal{R}$. $r=1$ signifies the successful completion of the task instruction.

In this work, we utilize ReAct \cite{yao2023react} to strengthen the reasoning ability of the LLM-based agent. Before taking an action, the LLM first generates a reasoning thought $t$. The trajectory $\tau$ will be transformed into:
\begin{equation*}
\tau = (u, t_1, a_1, o_1, ..., t_i, a_i, o_i, ..., t_n, a_n, o_n)
\end{equation*}

\begin{figure*}[htbp]
\centering
\includegraphics[width=\linewidth]{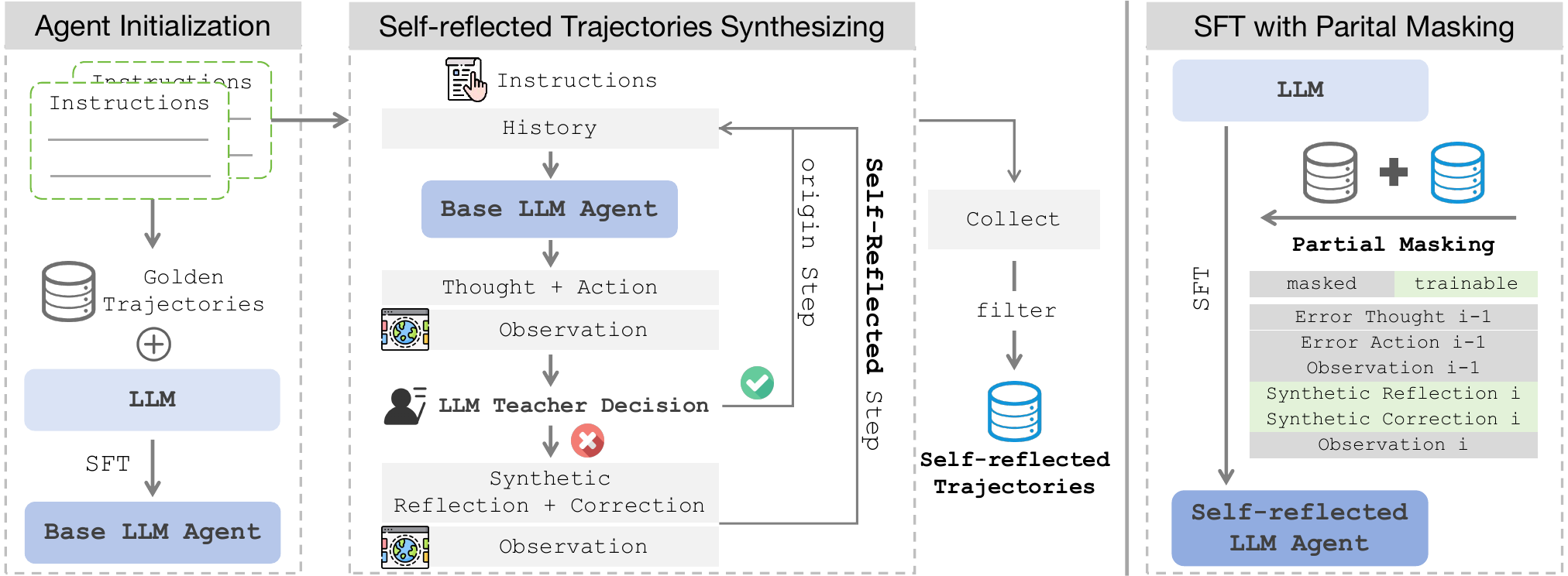}
\caption{STeP utilizes golden trajectories and corresponding instructions to train a Self-reflected LLM-based agent through three stages. Stage 1: Agent Initialization; Stage 2: \textit{Self-Reflected Trajectories} Synthesizing; Stage 3: SFT with \textit{Partial Masking}. }
\label{demo}
\end{figure*}

\subsection{Agent Initialization} \label{sec:stage1}
Following complex agent task instructions is quite challenging for smaller LLMs \cite{liu2024agentbench}, making it difficult for them to complete these tasks, even with the assistance of stronger LLMs. Moreover, even with in-context learning (ICL) \cite{brown2020language}, this issue remains hard to resolve. To obtain correctly formatted and effective trajectories, we first use a portion of expert interaction trajectory data to supervised fine-tune (SFT) the LLM $\pi_{base}$, creating a base LLM agent. The base LLM agent will act as a solid platform for developing more strong LLM-based agents.

Let $\mathcal{D}$ represent the set of all successful expert trajectories, which we refer to as golden trajectories. These trajectories contain interaction sequences that successfully complete each instruction in the task space $\mathcal{U}$. Each golden trajectory is organized in the ReAct format, encompassing reasoning thoughts and related actions.
We randomly select a subset of golden trajectories, designated as $\mathcal{D}_1$, corresponding to task subset $\mathcal{U}_1$. The remaining trajectories, denoted as $\mathcal{D}_2$, correspond to $\mathcal{U}_2$. We utilize $\mathcal{D}_1$ to supervised fine-tune an LLM, enabling it to acquire fundamental agent task knowledge and instruction-following capabilities. By minimizing the following loss function, we obtain the base LLM agent $\pi_{\theta}$ with parameters $\theta$:
\begin{align}
& \mathcal{L}_{SFT}(\theta) = -E_{(e, u,\tau) \sim \mathcal{D}_1} \left[ \log \pi_{\theta} (\tau | e, u) \right] \nonumber \\
& = -E_{(e, u,\tau) \sim \mathcal{D}_1} \left[ \sum_{i=1}^{n} \log \pi_{\theta} (t_i, a_i | e, u, \tau_{i-1}) \right]  \nonumber
\end{align}
where $\tau_{i-1}=(t_1, a_1, o_1, ..., t_{i-1}, a_{i-1}, o_{i-1})$.

\subsection{Synthesizing \textit{Self-Reflected Trajectories}} \label{sec:stage2}
Merely relying on learning golden trajectories from teacher LLMs limits the enhancement of the reflection mechanism in LLM-based agents and constrains further improvements in their agent capabilities. To address this issue, we propose \textit{Self-Reflected Trajectories}, which integrates reflections and corrections of failure steps within the trajectories. This approach aims to prevent error propagation and recurrence, thereby facilitating the successful completion of agent tasks.

For the agent tasks $\mathcal{U}_2$ in the remaining data $\mathcal{D}_2$, we let the base LLM agent $\pi_\theta$ interact with the environment $e$ corresponding to $u \in \mathcal{U}_2$ to generate trajectories. 
Concurrently, we introduce a more advanced LLM as the teacher model, denoted as $\pi_{teacher}$. $\pi_{teacher}$ will evaluate in real-time the correctness and rationality of actions taken by $\pi_\theta$ during its interactions with the environment $e$.
We design a complex prompt for the teacher model, which includes the definition and requirements of the agent task, the current instruction that the base LLM agent needs to complete, the interaction history $(t_1, a_1, o_1, ..., t_{i-1}, a_{i-1}, o_{i-1})$, the thought and action generated by $\pi_\theta$ in the present step $(t_i, a_i)$, and the corresponding observation $o_i$ given by the environment $e$.

After the base LLM agent $\pi_\theta$ outputs $(t_i, a_i)$, the teacher model $\pi_{teacher}$ judges whether it is correct. When an action is identified as incorrect, the teacher model will issue an error signal, marking that action and its corresponding thought as erroneous. 
Assume $\pi_\theta$ makes an error at step $j$, then $\pi_{teacher}$ will mark it as $(\widehat{t_j}, \widehat{a_j})$. Subsequently, based on the task requirements, the interaction history, and the feedback $o_j$ provided by $e$ at that step, $\pi_{teacher}$ will, in the first-person narrative, offer the content, reason, reflection on this error, and the correct action that $\pi_\theta$ should take to correct it. 
To ensure consistency in the training data, we convert the output of $\pi_{teacher}$ into the ReAct format and label it as $(t'_{j+1}, a'_{j+1})$. In this scenario, $t'_{j+1}$ includes the error itself along with the reflection on it, while $a'_{j+1}$ represents the action taken to correct the error. The environment will provide new feedback $o_{j+1}$ based on $a'_{j+1}$. If $\pi_{teacher}$ determines that the action of $\pi_\theta$ does not require correction, no operation will be performed. This process is repeated until the task is completed or the maximum step limit is reached.

In this way, we obtain $|\mathcal{U}_2|$ trajectories, which may contain error markers. We construct \textit{Self-Reflected Trajectories} set $\mathcal{D}_r$ by filtering out trajectories that successfully complete the agent task ($r = 1$) and include the teacher's reflections and correctness on the error steps. \textit{Self-Reflected Trajectory} $\tau'$ can be designating as:
\begin{equation*}
(u, t_1, a_1, o_1, ..., \widehat{t_j}, \widehat{a_j}, o_i, t'_{j+1}, a'_{j+1}, o_{j+1}, ...)
\end{equation*}

\begin{table}[b]
    \small
    \centering
\begin{tabular}{l|ccc}
\toprule
\textbf{Task} & Train & Test (seen) & Test (unseen) \\ \midrule
\textbf{ALFWorld} & 3119 & 140 & 134 \\
\textbf{SciWorld} & 1483 & 194 & 211 \\
\textbf{WebShop} & 1016 & 200 & - \\ \bottomrule
\end{tabular}
\caption{Statistics of the datasets. }
\label{data_stat}
\end{table}

\subsection{SFT with \textit{Partial Masking}} \label{sec:stage3}
Due to the tendency of catastrophic forgetting during supervised fine-tuning \cite{yang2024sdft}, where they may lose some of their original capabilities after training on downstream tasks, we employ a straightforward approach to mitigate this issue. Specifically, when training the final model with self-reflected trajectories, we do not base it on the base LLM agent $\pi_\theta$. Instead, we combine $\mathcal{D}_r$ with $\mathcal{D}_1$ to retrain the foundational LLM $\pi_{base}$.

Directly applying the SFT methods from the Agent stage may lead the LLM to learn error steps in the self-reflected trajectories, potentially impairing the agent's ability to take correct actions. Therefore, we introduce \textit{Partial Masking} (PM) to prevent the LLM from learning incorrect thoughts and actions. Specifically, during the training process, $(\widehat{t_j}, \widehat{a_j})$ will be masked, and the corresponding loss will not be computed. We let $\delta_i$ indicate whether step $i$ is an error step, $\delta_i=0$ indicates an error, while $\delta_i=1$ indicates otherwise.
The new loss function can be formulated as follows: 
\begin{equation*}
\begin{split}
\mathcal{L}_{PM}(\theta) &= -E_{(e, u,\tau) \sim {\mathcal{D}_1 + \mathcal{D}_r}} \left[ \log \pi_{\theta} (\tau | e, u) \right] \\
                         &= -E_{(e, u, \tau) \sim \mathcal{D}_1} \left[ \sum_{i=1}^{n} \right. \\
                         &\qquad \qquad \left. \log \pi_{\theta} (t_i, a_i | e, u, \tau_{i-1}) \right] \\ 
                         &\quad - E_{(e, u, \tau) \sim \mathcal{D}_r} \left[ \sum_{i=1}^{n} \right. \\
                         &\qquad \qquad \left. \delta_i \log \pi_{\theta} (t_i, a_i | e, u, \tau_{i-1}) \right]
\end{split}
\end{equation*}

\section{Experiments} \label{sec:exp}
\subsection{Datasets}
We perform our method on three agent tasks, namely ALFWorld, WebShop, and SciWorld. These tasks have been widely studied and are representative.
ALFWorld and SciWorld have a seen test set (within the distribution of the training set) and a unseen test set (outside the distribution), while WebShop only contains a seen test set. (See more information in Appendix \ref{app:data})

The datasets we used mainly follow ETO \cite{song2024eto}, which employed GPT-4 to generate ReAct thoughts for each step in tasks that inherently contain expert trajectories, or to generate expert trajectories for tasks that lack them. Due to some trajectories not fully completing the tasks, we filter out trajectories with $reward = 1$ to constitute our training set. We randomly sample 50\% of the data from each task and combine it to $\mathcal{D}_1$, which is used to train the base LLM agent during the Agent Initialization stage.

\subsection{Experiments Setup} \label{sec:setup}
Our experiments are primarily based on LLaMA2-7B-chat \cite{touvron2023llama2}, although we also conducted experiments with other models. Due to budget constraints, experiments with GPT-4 are not currently considered. Instead, we utilized several powerful open-source LLMs as teacher models and selected Qwen1.5-110B-Chat after testing. All LLMs are deployed using the vllm framework \cite{kwon2023vllm}. Given the challenges of reflection and correction, LLM teachers may also make errors; thus, we excluded self-reflected trajectories with a high number of erroneous steps. Ultimately, we obtained 708 \textit{Self-Reflected Trajectories}. During all inference phases, we set the temperature to 0 to ensure the determinism of the results. 

\begin{table*}[htbp]
    \small
    \centering
\begin{tabular}{c|l|c|cc|cc|cc}
\hline\hline
& \multirow{2}{*}{\textbf{LLM-based Agent}} & \multirow{2}{*}{\textbf{WebShop}} & \multicolumn{2}{c|}{\textbf{ALFWorld}} & \multicolumn{2}{c|}{\textbf{SciWorld}} & \multirow{2}{*}{\textbf{Average}} & \multirow{2}{*}{\textbf{Complate Rate}} \\
& &  & seen   & unseen  & seen   & unseen  &   &   \\ \midrule
\multirow{8}{*}{\rotatebox{90}{1-shot}}  & GPT-4o-0513  & 0.604 & 0.206 & 0.119 & 0.508 & 0.468 & 0.381 & 0.418 \\
& GPT-4          & 0.632 & 0.429 & 0.381 & 0.648 & 0.644 & 0.547 & — \\
& GPT-3.5-turbo  & 0.624 & 0.079 & 0.105 & 0.165 & 0.130 & 0.221 & — \\
\cmidrule{2-9}
& LLaMA3-70B-Instruct  & 0.589 & 0.234 & 0.252 & \textbf{0.699} & 0.618 & 0.478 & 0.524 \\
& Qwen1.5-110B-Chat    & 0.536 & 0.681 & \textbf{0.719} & 0.198 & 0.221 & 0.471 & 0.532 \\
& Qwen2-72B-Instruct   & 0.579 & 0.582 & 0.674 & 0.226 & 0.130 & 0.438 & 0.495 \\
\cmidrule{2-9}
& LLaMA3-8B-Instruct   & 0.562 & 0.035 & 0.037 & 0.333 & 0.324 & 0.258 & 0.384 \\
& LLaMA2-7B-Chat       & 0.185 & 0.000 & 0.000 & 0.028 & 0.025 & 0.048 & 0.076 \\ \midrule
\multirow{3}{*}{\rotatebox{90}{0-shot}} & AgentLM-7B   & \textbf{0.665} & 0.603 & 0.578 & — & — & — & — \\
& LLaMA2-7B-chat + Golden Trajs  & 0.565 & 0.688 & 0.652 & 0.65 & 0.611 & 0.636 & 0.707 \\  \cmidrule{2-9}
& LLaMA2-7B-chat + STeP (Ours)  & 0.618 & \textbf{0.716} & \textbf{0.719} & 0.664 & \textbf{0.660} & \textbf{0.672} & \textbf{0.754} \\ \hline\hline
\end{tabular}
\caption{The main results of our experiments. Values in the columns for WebShop, ALFWorld, and SciWorld represent the \textbf{average reward} for each model on the respective tasks (see Section \ref{sec:exp}). The ``Average" column represents the mean of the values from the first five columns. The final column indicates the completion rate of all tasks (the ratio of tasks completed before reaching the maximum step limits or the maximum context length).}
\label{results}
\end{table*}

\subsection{Baselines}
We evaluate other LLMs for comparison. 
1). Golden trajectories only: \textbf{FireAct} \cite{chen2023fireact} and \textbf{AgentTuning} \cite{agenttuning} utilize expert trajectories only to SFT LLMs. We employ the golden trajectories $\mathcal{D}$ only to SFT LLaMA2-7B-chat, referring to as LLaMA2-7B-chat + Golden Trajs. We also tested AgentLM-7B from AgentTuning, which includes general datasets in its training data. Since its training set only covers ALFWorld and WebShop, we only evaluate it on these two datasets.
2). Base LLMs that have not been trained on these agent tasks: GPT-4o-0513, GPT-4, GPT-3.5-turbo, LLaMA3-70B-Instruct \cite{llama3}, Qwen1.5-110B-Chat, Qwen2-72B-Instruct \cite{yang2024qwen2technicalreport}, LLaMA2-7B-Chat, and LLaMA3-8B-Instruct.

\subsection{Evaluation}
The performance of LLM-based agents can be represented by reward $r \in [0, 1]$, where $r=1$ signifies complete success. The reward $r$ is automatically derived from the environment $e$ corresponding to the agent task. The main evaluation metric is \textbf{average reward}, which is the mean of the rewards the agent obtains across all instructions in a certain agent task. Reward calculation varies by task. In ALFWorld, reward 1 corresponding to task completion and 0 to failure. For the other tasks, the reward reflects the completion level of the task.

Agent tasks have maximum step limits. If the agent fails to complete the task instruction within the given steps, the reward $r=0$. (Details can be seen in Appendix \ref{app:eval}). Due to limited context window, we test LLM-based agent in zero-shot settings.
However, since base LLMs have not been specifically trained on agent tasks, the zero-shot setting presents challenges. Consequently, we adopted a one-shot approach for these LLMs.

\section{Results and Analysis}
Based on the experimental settings in Section \ref{sec:exp}, we test our trained LLM-based agent along with various baselines on three agent tasks, as presented in Table~\ref{results}. We also perform ablation experiments and provide a detailed analysis in this section.

\subsection{Main Results}
\begin{figure}[t]
\centering
\includegraphics[width=1\columnwidth]{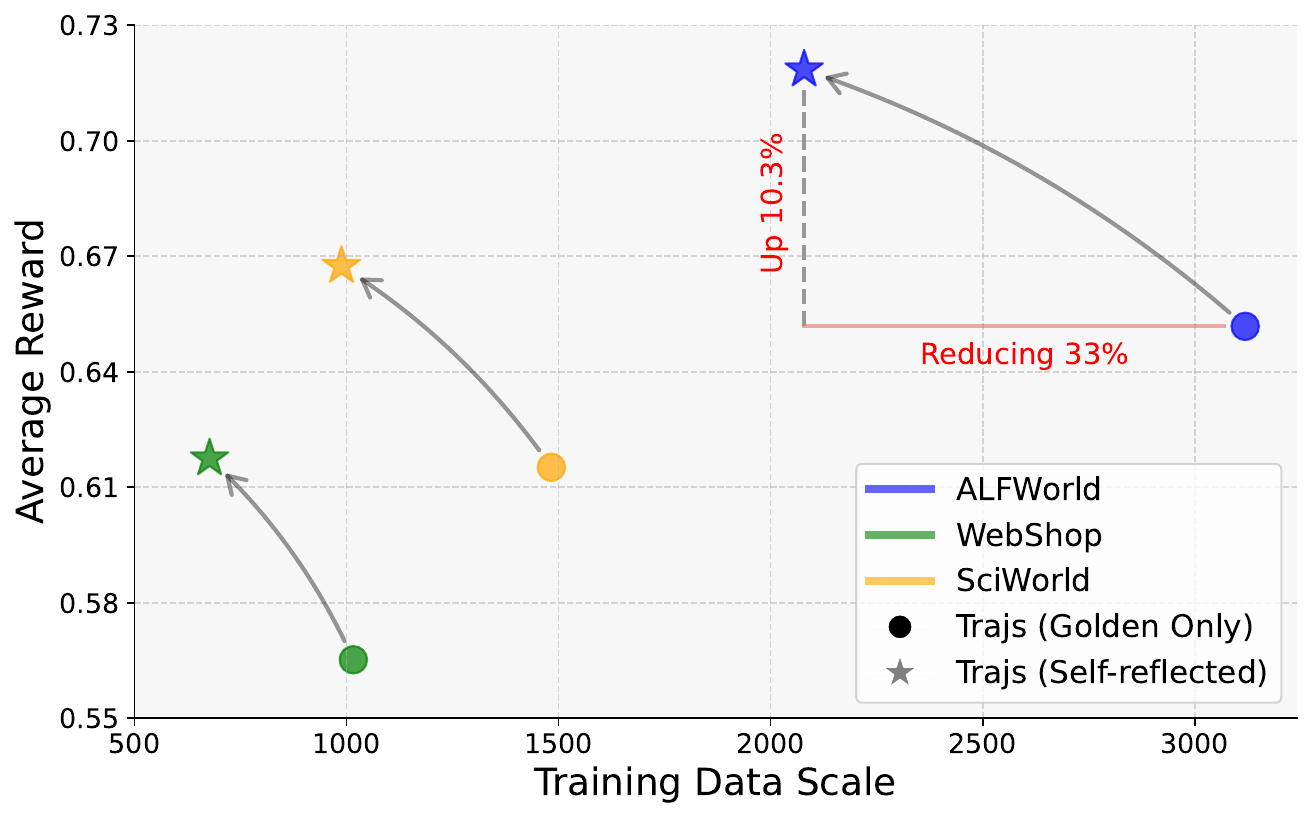} 
\caption{Compared to golden only, self-reflected trajectories help LLMs learn more effectively and efficiently.}
\label{compare}
\end{figure}

Our agent outperforms most baselines in average reward across all tasks, demonstrating the efficacy of STeP.
Compared to agents trained solely on golden trajectories, STeP achieves better results across all test sets. LLaMA2-7B-chat + STeP outperforms LLaMA2-7B-chat + Golden Trajs by an average of 9.2\% on all tasks. Specifically, it improved performance by 9.4\% on WebShop, 4.1\% and 10.3\% on ALFWorld (seen / unseen), 2.2\% and 8\% on SciWorld (seen / unseen). 
Moreover, these improvements are achieved with fewer self-reflected trajectories than golden trajectories, showing that reflection-based training data helps LLMs learn more effectively and efficiently (Figure \ref{compare}). Improving the quality and efficiency of self-reflected trajectories will further enhance our results.

As illustrated in Table \ref{results}, LLMs that have not been specifically trained on these agent tasks struggle to complete them, particularly smaller LLMs. LLaMA3-8B-Instruct and LLaMA2-7B-chat struggle with agent tasks like ALFWorld. Using trajectories for SFT improves performance, achieving at least 50\% reward and approaching or surpassing larger closed-source LLMs. The agent trained by STeP consistently outperforms its teacher models.
Larger closed-source LLMs average about 0.6 reward on WebShop. Besides, Qwen excels in ALFWorld and others in SciWorld, likely due to differences in their training data. Some open-source LLMs may benefit from exposure to agent-specific datasets. Furthermore, although adding reflection and error correction may increase the number of steps and tokens to some extent, our method still achieves an improvement in the completion rate.

\begin{table*}[t]
    \small
    \centering
\begin{tabular}{l|c|cc|cc|cc}
\hline\hline
\multirow{2}{*}{\textbf{LLM-based Agent}} & \multirow{2}{*}{\textbf{WebShop}} & \multicolumn{2}{c|}{\textbf{ALFWorld}} & \multicolumn{2}{c|}{\textbf{SciWorld}} & \multirow{2}{*}{\textbf{Average}} & \multirow{2}{*}{\textbf{Complate Rate}} \\
&  & seen   & unseen  & seen   & unseen  &   &   \\ \midrule
LLaMA2-7B-chat + STeP  & \textbf{0.618} & 0.716 & \textbf{0.719} & \textbf{0.664} & \textbf{0.660} & \textbf{0.672} & \textbf{0.754} \\
\quad \quad \quad \textit{$-$ partial masking}   & 0.595 & \textbf{0.816} & \textbf{0.719} & 0.620 & 0.576 & 0.665 & 0.753 \\ 
\quad \quad \quad \textit{$-$ self-reflected trajectories}   & 0.555 & 0.702 & 0.667 & 0.656 & 0.541 & 0.624 & 0.697 \\ \hline\hline
\end{tabular}
\caption{Without using \textit{Self-Reflected Trajectories} and \textit{Partial Masking}, the overall performance of the agent decline.}
\label{ablation}
\end{table*}

\begin{figure}[t]
\centering
\includegraphics[width=1\columnwidth]{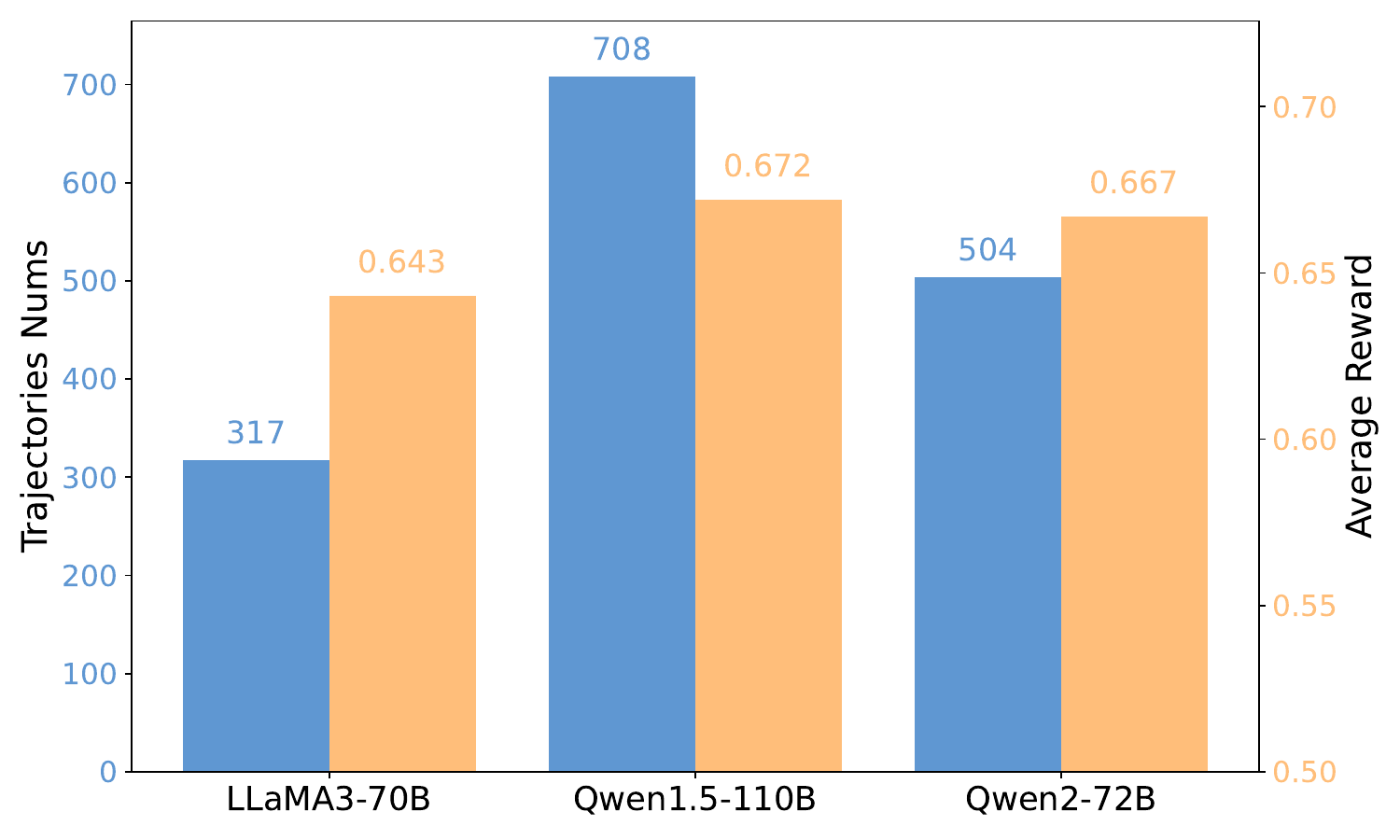} 
\caption{The number of Self-Reflected Trajectories generated by different teacher models, along with the average reward of the LLM-based agent trained on them.}
\label{teacher}
\end{figure}

\begin{table}[t]
    \small
    \centering
\begin{tabular}{l|cccc}
\toprule
\textbf{LLM-based Agent} & \textbf{Web} & \textbf{Alf} & \textbf{Sci} & \textbf{Avg} \\ \midrule
Mistral-7B + SFT & 0.406 & 0.681 & 0.661 & 0.583 \\
Mistral-7B + STeP & 0.454 & 0.741 & 0.571 & \textbf{0.589} \\ \midrule
NAT & 0.616 & 0.597 & 0.488 & 0.567 \\
WKM & 0.664 & 0.659 & 0.548 & 0.624 \\
LLaMA3-8B + SFT & 0.604 & 0.704 & 0.664 & 0.657 \\
LLaMA3-8B + STeP & 0.618 & 0.763 & 0.630 & \textbf{0.670} \\ \bottomrule
\end{tabular}
\caption{With STeP, agents trained based on other LLMs all achieve overall superior performance.}
\label{other}
\end{table}

\begin{table}[t]
    \small
    \centering
\begin{tabular}{l|ccc}
\toprule
\textbf{LLM-based Agent} & \textbf{Web} & \textbf{Alf} & \textbf{Sci} \\ \midrule
LLaMA2-7B + STeP & \textbf{0.618} & 0.719 & \textbf{0.660} \\
LLaMA2-7B (Step) & 0.563 & 0.770 & 0.550 \\ \midrule
1-Shot: & & & \\
LLaMA2-7B + STeP & 0.405 & 0.674 & 0.551 \\
LLaMA2-7B + Golden Trajs & 0.434 & 0.630 & 0.546 \\ \midrule
LLaMA2-7B (WebShop)   & 0.605 & — & — \\
LLaMA2-7B (ALFWorld)  & — & \textbf{0.822} & — \\
LLaMA2-7B (SciWorld)  & — & — & 0.596 \\ \midrule
LLaMA3-70B + self-ref & 0.262 & 0.237 & 0.441 \\
Qwen1.5-110B + self-ref & 0.429 & 0.793 & 0.227 \\ \bottomrule
\end{tabular}
\caption{\textbf{1.} Combining Self-Reflected Trajectories with partial golden trajectories for retraining the LLM helps mitigate catastrophic forgetting. \textbf{2.} Due to the limited context window, the 1-shot performance is suboptimal. \textbf{3.} Training on a single task does not necessarily enhance the model's performance on the corresponding task. \textbf{4.} Finally, STeP not only surpasses the simple imitation of correct behaviors from the teacher LLM but also outperforms the self-reflection of the teacher LLM itself.}
\label{other2}
\end{table}

\subsection{Ablation Experiments}
\paragraph{The Effectiveness of Self-Reflected Trajectories.}
Our training set combines \textit{Self-Reflected Trajectories} $\mathcal{D}_r$ with a portion of golden trajectories $\mathcal{D}_1$. To evaluate the impact of the self-reflected trajectories, we trained LLaMA2-7B-chat using only $\mathcal{D}_1$, resulting in the model LLaMA2-7B-chat (\textit{w/o self-reflected trajectories}). As illustrated in Table \ref{ablation}, not utilizing any self-reflected trajectories leads to a decrease in performance.
If increasing the number of golden trajectories and training with $\mathcal{D}$, i.e. LLaMA2-7B-chat + Golden Trajs, the gains are modest: a 12.9\% gain on the unseen SciWorld test set, 1.8\% on WebShop, and even a 2.2\% decline on ALFWorld unseen test set. In contrast, our method added only 708 self-reflected trajectories to $\mathcal{D}_1$, resulting in improvements of 22\% (SciWorld), 11.4\% (WebShop) and 7.8\% (ALFWorld), the average increase is 13.7\%.

\paragraph{The Effectiveness of Partial Masking.}
We replaced the loss function from Section \ref{sec:stage3} with the one from Section \ref{sec:stage1}, enabling the model to learn from actions and thoughts labeled as incorrect. The results indicated \textit{Partial Masking} effectively boosts the overall agent capabilities of the LLM. However, the agent trained without the mask achieved better results on the ALFWorld seen test set, surpassing all agents. We suggest that this may be because this tasks have a higher tolerance for incorrect actions, allowing the model without the partial mask to explore a broader range of erroneous trajectories. We plan to explore this in future works.

\subsection{Analysis}
\paragraph{The Selection of The LLM Teacher.}
We conducted experiments with several open-source LLMs to identify the more suitable LLM teacher (see Appendix \ref{app:selection} for detailed information).
As depicted in Figure \ref{teacher}, it is evident that Qwen1.5-110B-Chat produced the most \textit{Self-Reflected Trajectories} and yielded the best test results when used as the teacher model. Although other LLM teachers did not perform as well as Qwen1.5-110B-Chat, they still achieved notable results. The LLM-based agents trained with these other LLM teachers outperformed LLaMA2-7B-chat + Golden Trajs.

\paragraph{Experiments Based on Other LLMs.} 
We use \textit{Self-Reflected Trajectories} $\mathcal{D}_r$ alongside the half of all golden trajectories $\mathcal{D}_1$ to train Mistral-7B-Instruct-V0.3 \cite{jiang2023mistral7b} and LLaMA3-8B-Instruct according to our proposed method. 
In addition, we compared two other methods based on LLaMA3-8B: WKM \cite{wkm}, which introduced a parametric world knowledge model to facilitate agent planning, and NAT \cite{wang2024nat}.
The experimental results indicate that while both models perform below the corresponding SFT results on the SciWorld task, they surpass those results on the WebShop and ALFWorld tasks, demonstrating an overall superior performance.

\paragraph{The Effectiveness of Combine $\mathcal{D}_r$ with $\mathcal{D}_1$.} We combine $\mathcal{D}_r$ with $\mathcal{D}_1$ to retrain the foundational LLM $\pi_{base}$ to mitigate the catastrophic forgetting problem. To verify this, we design an experiment in which we first train on $\mathcal{D}_1$ and then conduct further training on $\mathcal{D}_r$. We named the resulting agent LLaMA2-7B (Step). The results show improved efficacy on the ALFWorld task, while results on other tasks declined significantly.
The results in  Table \ref{other2} indicate that a better outcome was achieved on the ALFWorld task, while results on other tasks declined significantly. This validates that combining $\mathcal{D}_r$ with $\mathcal{D}_1$ is necessary to elevate the overall performance of the agent.

\paragraph{Zero-shot or One-shot.} The main results of our agents are based on zero-shot prompting. Here, we conduct one-shot experiments to explore the impact of in-context learning (ICL) \cite{brown2020language} examples on our approach. Compared to zero-shot prompting, using one-shot for inference significantly affects model agent abilities. We hypothesize that this is due to limitations in the context window and the potential disparity between the ICL examples and the target instructions.

\paragraph{Training on Single Task or Multiple Tasks.} We use the golden trajectories from three agent tasks to train LLaMA2-7B-Chat individually. As reflected in the results presented in Tables \ref{other} and \ref{results}, training on a single task only benefits certain agent tasks, while for WebShop and SciWorld, training on multiple tasks yields better results.

\paragraph{Comparison with Teacher LLMs' Self-reflection.} We followed the procedure outlined in Section \ref{sec:stage2}, allowing the teacher LLM to engage in self-reflection during the testing process (denoted as LLM + self-ref). As shown in Table \ref{other2}, STeP not only surpasses the simple imitation of correct behaviors from the teacher LLM, but also outperforms the self-reflection of the teacher LLM itself. This indicates that, to some extent, we do not rely solely on the capabilities of the teacher.

\section{Conclusion}
In this paper, we introduce STeP, a novel method that leverages \textit{Self-Reflected Trajectories} to enable smaller LLMs to learn how to generate reflections and corrections of error steps and employs \textit{Partial Masking} to help LLMs avoid generating erroneous thoughts and actions. STeP could make the process of learning agent capabilities from teacher models more effective and efficient. Experiments on several representative agent task datasets demonstrate the validity of our approach. We hope STeP could inspire future researchers to create more capable LLM-based agents.

\section*{Limitations}
Our method still has limitations. 
For instance, the current success rate and effectiveness of using prompts to enable the LLM teacher to reflect and correct in real-time need further improvement. Reflections and corrections of erroneous steps in self-reflected trajectories may not always be appropriate. We aim to address these issues in future works.
Secondly, due to small-scale LLMs have relatively limited capabilities and are unable to generate sufficient high-quality trajectories. Therefore, utilizing the base LLM agent itself as a teacher model remains challenging, necessitating the assistance of more powerful LLMs. However, we suggest that to enable LLMs to develop self-reflect capabilities from data, the key lies not in whether to use another teacher model, but rather in how to effectively learn from the teacher LLM. We will also attempt to conduct experiments using self-feedback from the base LLM in the future.

\bibliography{custom}

\appendix

\section{Datasets and Prompts} \label{app:data}
\begin{figure*}[h]
\centering
\includegraphics[width=2.1\columnwidth]{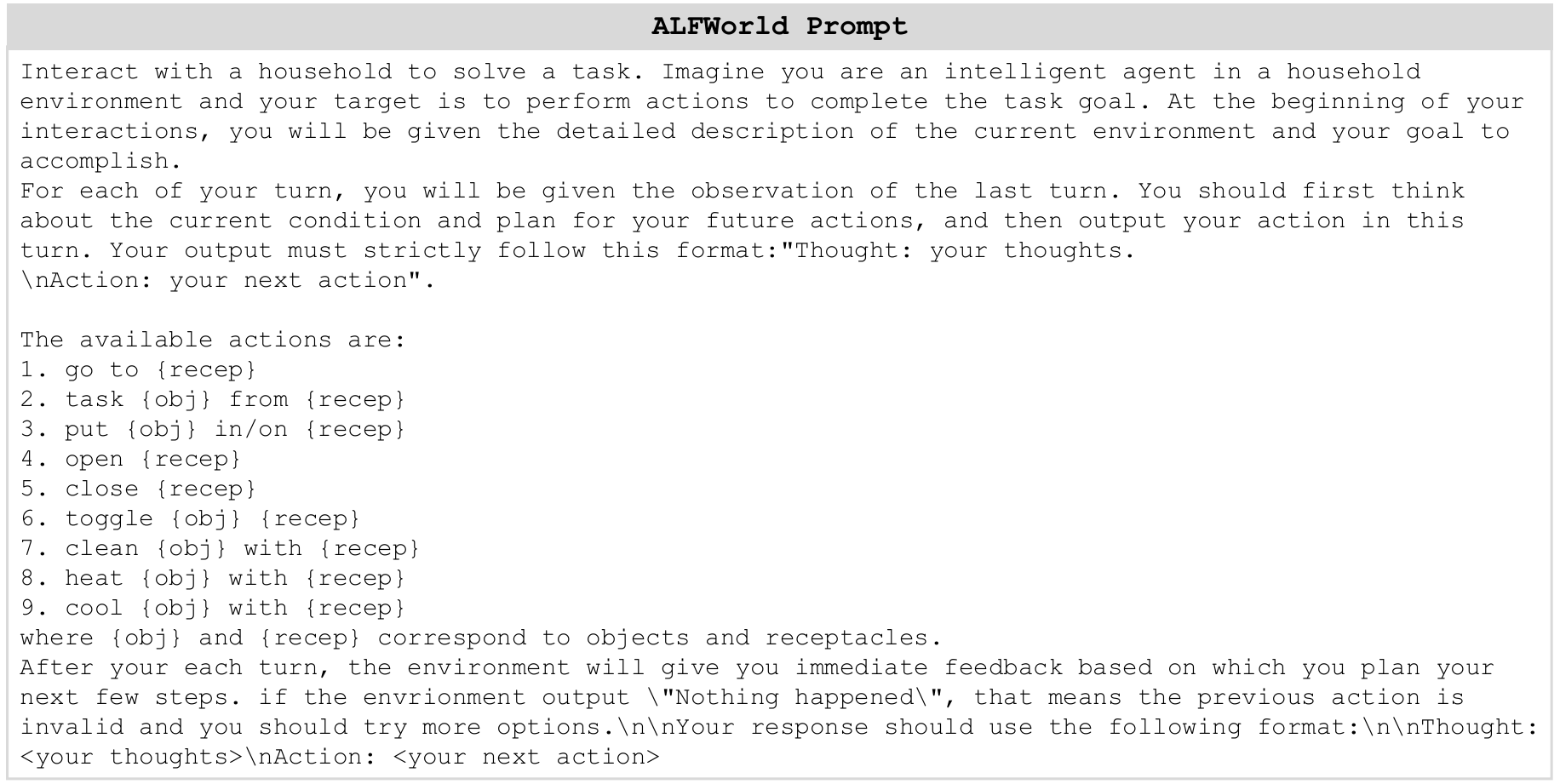} 
\caption{The prompt of ALFWorld that contains task requirements.}
\label{alfworld}
\end{figure*}

\begin{figure*}[h]
\centering
\includegraphics[width=2.1\columnwidth]{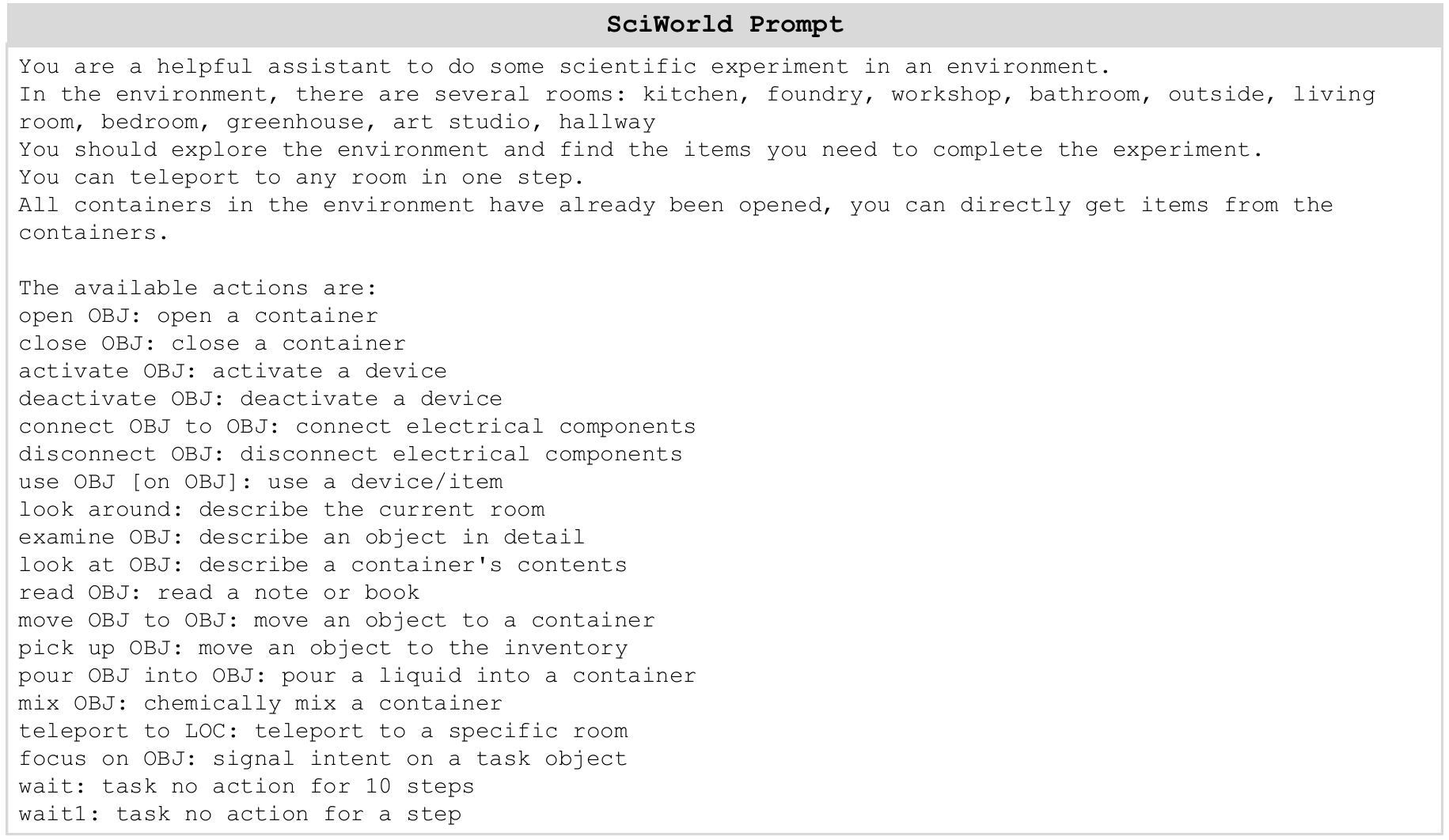} 
\caption{The prompt of SciWorld that contains task requirements.}
\label{sciworld}
\end{figure*}

\begin{figure*}[h]
\centering
\includegraphics[width=2.1\columnwidth]{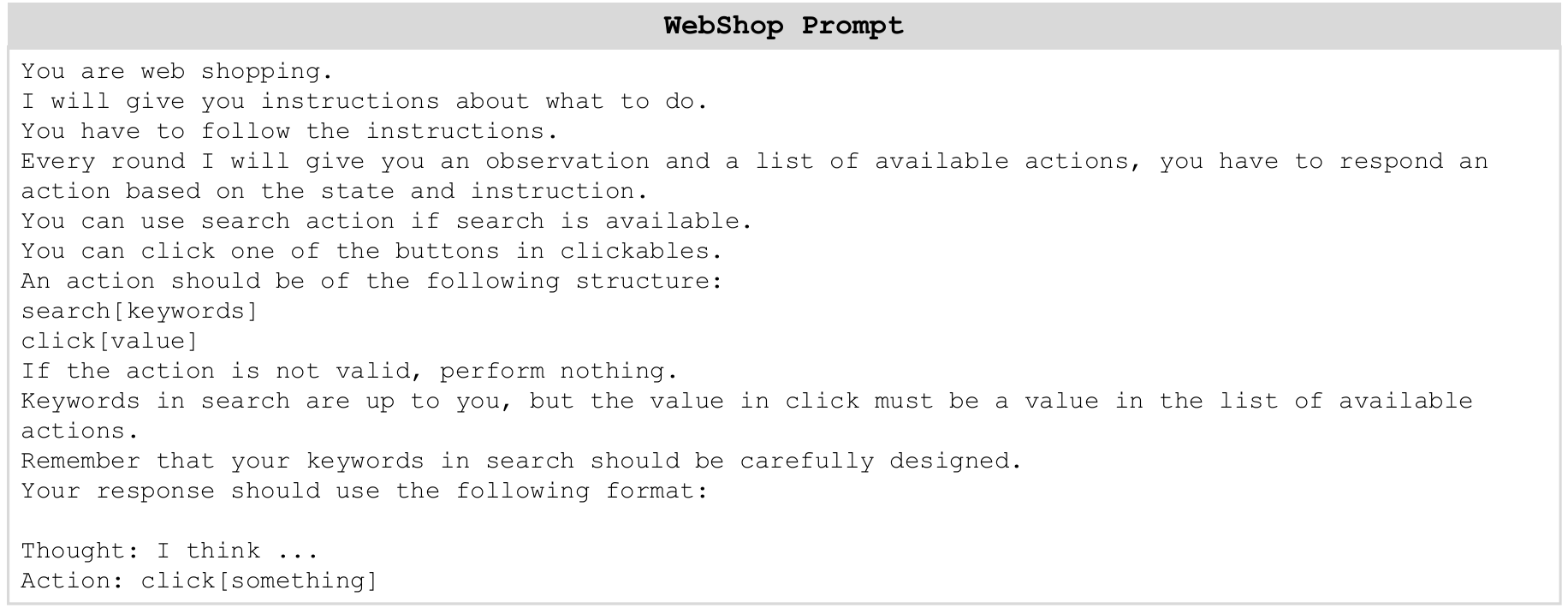} 
\caption{The prompt of WebShop that contains task requirements.}
\label{webshop}
\end{figure*}

We experiment on 3 representative agent tasks:
\begin{itemize}
\item \textbf{ALFWorld}: ALFWorld contains text-simulated environments that parallel embodied worlds in the ALFRED dataset \cite{Shridhar_2020_CVPR}. The agent must provide textual actions to solve simulated embodied tasks, such as placing a vase in a safe. Specific task formats and requirements are detailed in Figure \ref{alfworld}. These task requirements will be used as prompts for each LLM agent input into the model.
\item \textbf{SciWorld}: SciWorld aims to test agents' scientific reasoning skills within an interactive text environment, corresponding to a standard elementary school science curriculum. Specific task formats and requirements are illustrated in Figure \ref{sciworld}.
\item \textbf{WebShop}: In WebShop tasks, the LLM-based agent is required to purchase products according to the user instructions and the web information. The agent must simulate actions like searching, clicking, and purchasing using text, as shown in Figure \ref{webshop}.
\end{itemize}

\section{Experiments Setup}
\subsection{Hyper Parameters}
In the Agent Initialization and SFT with \textit{Partial Masking} stages, we employ supervised fine-tuning (SFT) to train the LLM, utilizing the AdamW optimizer. The batch size is set to 32, and the learning rate is configured to 3e-5 with a cosine scheduler. The learning epochs for all SFT phases are set to 4. All training experiments are conducted on 4 NVIDIA A100 80G GPUs. We primarily conducted the fine-tuning process of LLMs using the LLaMA-Factory \cite{zheng2024llamafactory} framework. We made certain modifications to the data processing and model training components to ensure the execution of \textit{Partial Masking}.

\subsection{Model Deployment}
In the Trajectory Augmentation stage, for closed-source LLM teachers, we access closed-source LLMs via API, while open-source LLM teachers are deployed using the vllm framework \cite{kwon2023vllm} on NVIDIA A100 80G GPUs. In this stage, we need to execute inference for the base LLM agent. We also use vllm to deploy it on a single NVIDIA GTX 4090 24G GPU, setting the temperature to 0 to ensure the determinism of the results.

\subsection{Self-reflected Trajectories Construction} \label{sec:setup}
Using the LLM teacher for real-time reflection and correction would consume a significant number of tokens. Due to budget constraints, experiments with GPT-4 are not currently considered. Instead, we use several powerful open-source LLMs as teacher models, including LLaMA3-70B-Instruct \cite{llama3}, Qwen1.5-110B-Chat, and Qwen2-72B-Instruct \cite{yang2024qwen2technicalreport}. LLM teachers and the base LLM agent are deployed using the vllm framework \cite{kwon2023vllm}. During all inference phases, we set the temperature to 0 to ensure the determinism of the results. Due to the difficulty of the reflection and correction, LLM teachers may also make errors. Therefore, we excluded self-reflected trajectories with a high number of erroneous steps. Specifically, for ALFWorld and SciWorld, a maximum of 2 erroneous steps is allowed, while for Webshop, the limit is 1. Ultimately, we selected Qwen1.5-110B-Chat as the teacher model, resulting in 708 \textit{Self-Reflected Trajectories}. 

\subsection{Evaluation Details} \label{app:eval}
The performance of LLM-based agents in completing a given task instruction can be represented by reward $r \in [0, 1]$, where $r=1$ signifies complete success. The reward $r$ is automatically derived from the environment $e$ corresponding to the agent task. The main evaluation metric is \textbf{average reward}, which is the mean of the rewards the agent obtains across all instructions in a certain agent task. Reward calculation varies by task. In ALFWorld, reward 1 corresponding to task completion and 0 to failure. For the other tasks, the reward reflects the completion level of the task.

Agent tasks have maximum step limits. If the agent fails to complete the task instruction within the given steps, the reward $r=0$. Specifically, the maximum steps for ALFWorld tasks is 30, for WebShop it is 10, and for different subtasks in SciWorld, the step limits vary from 10 to 120. If the number of tokens in the trajectories exceeds the maximum context length of the LLM, $r$ is also set to 0.

During testing, we set the temperature of all models to 0, using greedy decoding to obtain deterministic results. We primarily utilize a zero-shot prompt to guide the LLM-based agent in completing task instructions step-by-step in the ReAct format. This is because of the limited context window length and the fact that the LLM has been trained on trajectories adhering to this format, allowing it to follow task requirements to some extent. Additionally, we conducted one-shot experiments for comparative analysis. 
However, since base LLMs have not been specifically trained on agent tasks, the zero-shot setting presents challenges. Consequently, we adopted a one-shot approach when testing these LLMs.

\subsection{Teacher Model Setup}
\begin{figure*}[h]
\centering
\includegraphics[width=2.1\columnwidth]{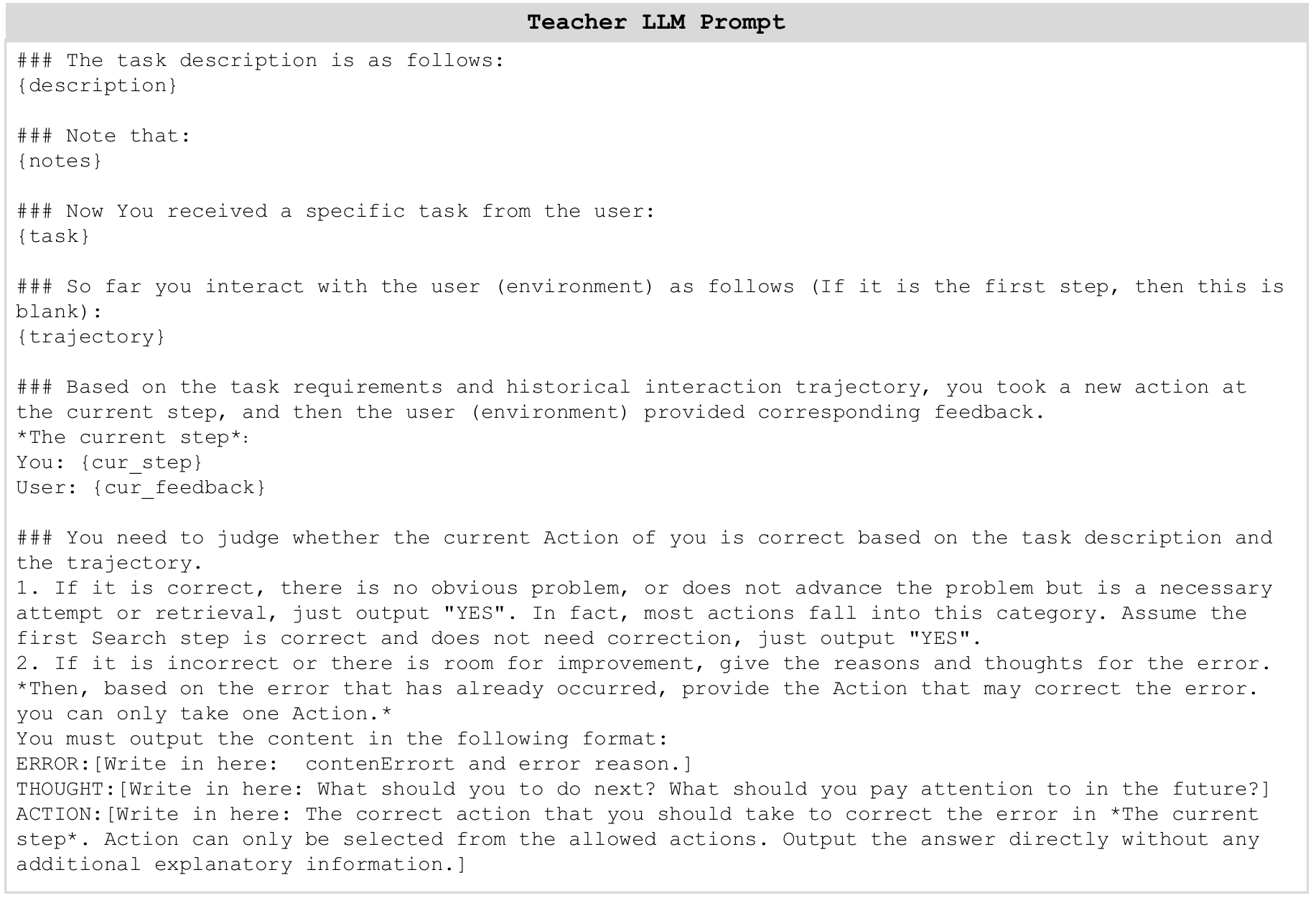} 
\caption{The prompt for teacher models to reflect and correct in real-time.}
\label{appendix}
\end{figure*}

We utilize a teacher model to evaluate in real-time whether the actions taken by the base LLM agent are correct. To facilitate this, we designed prompts (Figure \ref{appendix}) that include specific instructions for each agent task, as shown in Figures \ref{alfworld} to \ref{webshop}. Additionally, to assist the LLM teacher in making better decisions, we created several corresponding considerations based on the characteristics of each task.
The LLM teacher will combine the current task instruction, interaction history, the actions taken by the base LLM agent, and the corresponding feedback from the environment to determine whether the action is correct. If the action is correct, the teacher will simply output "yes" without any modifications. If incorrect, the teacher will provide the error itself, the reason for the error, reflections on the mistake, and how to act to resolve the issue. The ERROR and THOUGHT components in the prompt will be treated as the thought process in the ReAct format. To ensure consistency, the teacher model is instructed to output in the first-person narrative.

\subsection{The Selection of LLM Teachers}  \label{app:selection}
We conducted experiments with several open-source LLMs to identify the more suitable LLM teacher. 
In the Trajectory Augmented stage, we utilized LLaMA3-70B-Instruct, Qwen1.5-110B-Chat, and Qwen2-72B-Instruct as LLM teachers to provide real-time reflection and correction for LLaMA2-7B-chat. 
For each LLM teacher, we filtered the corresponding \textit{Self-Reflected Trajectories} based on the methods outlined in Section \ref{sec:stage3} and Section \ref{sec:setup}. Subsequently, using these data, we trained the corresponding models according to the method described in Section \ref{sec:stage3} and ultimately obtained the average reward of these models across all test sets.

\end{document}